\pdfoutput=1

\documentclass[11pt]{article}

\usepackage{times}
\usepackage{latexsym}
\usepackage[T1]{fontenc}
\usepackage[utf8]{inputenc}
\usepackage{microtype}
\usepackage{inconsolata}
\usepackage{graphicx}

\usepackage[final]{acl}
\usepackage{times}
\usepackage{latexsym}

\usepackage[T1]{fontenc}

\usepackage[utf8]{inputenc}

\usepackage{microtype}

\usepackage{inconsolata}

\usepackage{graphicx}

\title{Zero-Shot Commonsense Validation and Reasoning with Large Language Models: An Evaluation on SemEval-2020 Task 4 Dataset}

\author{Rawand Alfugaha \\
    College of Information Technology \\
  Lusail University \\
  Doha, Qatar \\
  \texttt{ralfoqha@lu.edu.qa} \\
  \And
  Mohammad AL-Smadi \\
 Digital Learning and Online Education Office  \\
 Qatar University\\  Doha, Qatar \\
 \texttt{malsmadi@qu.edu.qa} \\
}

\begin{document}

\maketitle
\begin{abstract}
 This study evaluates the performance of Large Language Models (LLMs) on SemEval-2020 Task 4 dataset, focusing on commonsense validation and explanation. Our methodology involves evaluating multiple LLMs, including LLaMA3-70B, Gemma2-9B, and Mixtral-8x7B, using zero-shot prompting techniques. The models are tested on two tasks: Task A (Commonsense Validation), where models determine whether a statement aligns with commonsense knowledge, and Task B (Commonsense Explanation), where models identify the reasoning behind implausible statements. Performance is assessed based on accuracy, and results are compared to fine-tuned transformer-based models. The results indicate that larger models outperform previous models and perform closely to human evaluation for Task A, with LLaMA3-70B achieving the highest accuracy of \textbf{98.40\%} in Task A whereas, lagging behind previous models with 	\textbf{93.40\%} in Task B. However, while models effectively identify implausible statements, they face challenges in selecting the most relevant explanation, highlighting limitations in causal and inferential reasoning. 

\end{abstract}

\section{Introduction}
Commonsense reasoning is a crucial aspect of Natural Language Processing (NLP) that enables models to understand and validate knowledge beyond explicit textual data. The motivation behind this research comes from the need to develop NLP models that can reason beyond surface-level text representations and apply real-world knowledge to language understanding tasks. Existing benchmarks, such as CommonGen \cite{lin2020commongen}, SemEval-2020 Task 4: Commonsense Validation and Explanation \cite{wang2020semeval}, CommonSenseQA 2.0 \cite{talmor2021commonsenseqa}, and COPEN \cite{peng2023copen}, have highlighted various aspects of commonsense reasoning, including generative commonsense reasoning, multi-hop reasoning, and physical commonsense knowledge. However, these tasks still pose challenges in handling nuanced reasoning \cite{10.1007/3-540-45757-7_24}, causal inference\cite{yao2021survey}, and knowledge integration \cite{chen2020review}.

The SemEval-2020 Task 4: Commonsense Validation and Explanation \cite{wang2020semeval} has served as a benchmark for evaluating various NLP models' capabilities in this domain. The task consistes of three sub-tasks, where in this research we will focus on the first two namely: Task A - Commonsense Validation: Determining whether a given statement aligns with commonsense knowledge, and Task B - Commonsense Explanation:	Identifying the reasoning behind why a statement is against common sense. Table~\ref{tab:commonsense_tasks} provides examples on both tasks as they appear in the dataset.

\begin{table*}[h]
    \centering
    \begin{tabular}{|l|p{9cm}|}
        \hline
        \textbf{Task} & \textbf{Example} \\
        \hline
        \textbf{Task A: Commonsense Validation} & 
        \textbf{Which statement is against common sense?} \newline
        - \textbf{Statement 1:} He put a turkey into the fridge. ( Correct) \newline
        - \textbf{Statement 2:} He put an elephant into the fridge. (Against commonsense) \\
        \hline
        \textbf{Task B: Commonsense Explanation} & 
        \textbf{Why is this statement against common sense?} \newline
        \textbf{Statement:} He put an elephant into the fridge. \newline
        - \textbf{A:} An elephant is much bigger than a fridge. ( Correct) \newline
        - \textbf{B:} Elephants are usually white while fridges are usually white. \newline
        - \textbf{C:} An elephant cannot eat a fridge. \\
        \hline
    \end{tabular}
    \caption{Examples of Commonsense Validation and Explanation Tasks}
    \label{tab:commonsense_tasks}
\end{table*}

This paper aims to explore how well large language models (LLMs) perform on commonsense reasoning tasks using zero-shot prompting. By evaluating multiple LLMs on SemEval-2020 Task 4, we investigate their ability to reason effectively without explicit fine-tuning. We present an overview of existing research, detail our methodology, and analyze experimental results to assess the strengths and limitations of current approaches.

\section{Related Work}

 SemEval-2020 Task 4, which focuses on Commonsense Validation and Explanation, attracted considerable attention, with numerous teams participating in its three subtasks. This literature review highlights the best-performing models in Tasks A and B, showcasing their methodologies and contributions to the field.

CN-HIT-IT.NLP \cite{zhang2020cn} emerged as the leading model in Subtask A, employing a variant of K-BERT  \cite{liu2019kbert} as its encoder. This model stands out for its integration of knowledge graphs, specifically ConceptNet \cite{10.5555/3298023.3298212}, which allows it to extract relevant triples that enhance the understanding of language representations. This approach underscores the importance of leveraging structured knowledge to improve commonsense reasoning capabilities.

In Subtask B, ECNU-SenseMaker \cite{zhao2020ecnu} achieved top performance by also utilizing K-BERT \cite{liu2019kbert}. This model innovatively combines structured knowledge from ConceptNet \cite{10.5555/3298023.3298212} with unstructured text through a Knowledge-enhanced Graph Attention Network. This integration facilitates a deeper understanding of commonsense knowledge, demonstrating the effectiveness of combining different types of information to enhance model performance.

Another notable model, IIE-NLP-NUT \cite{xing-etal-2020-iie}, utilized RoBERTa as its encoder. This model's unique contribution lies in its second pretraining phase, which involved a textual corpus from the Open Mind Common Sense (OMCS) project \cite{10.1007/3-540-36124-3_77}. By exploring various prompt templates for input construction, this model illustrates the potential of tailored input strategies in improving commonsense validation tasks

Team Solomon \cite{srivastava-etal-2020-team} was ranked 5th and 4th in Subtasks A and B, respectively. Their approach, which also relied on RoBERTa, highlighted the capacity of large-scale pretrained language models to encapsulate commonsense knowledge effectively without external resources. 

Across the two subtasks, the dominant trend was the use of large-scale pretrained language models such as  K-BERT  \cite{liu2019kbert}, \textit{RoBERTa} \cite{liu2019roberta}, \textit{BERT} \cite{devlin2018bert}, and \textit{GPT-2} \cite{radford2019gpt2}, often fine-tuned with additional commonsense knowledge sources. Additionally, models incorporating external structured knowledge sources (e.g., ConceptNet) generally outperformed purely language-model-based approaches.
\begin{figure*}[t]
    \centering
  \includegraphics[width=1.0\linewidth]{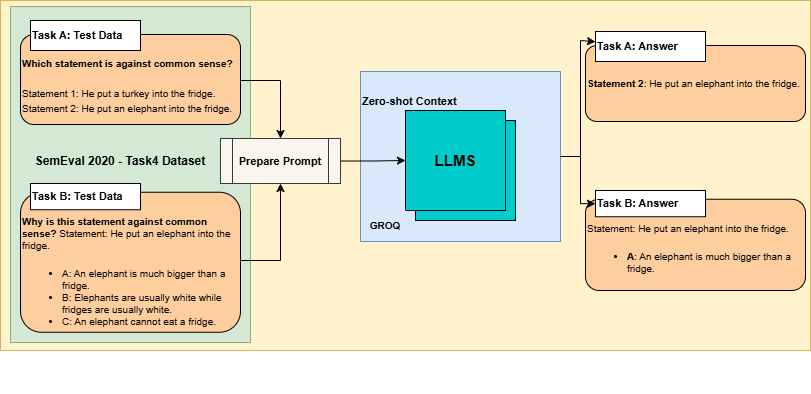} \hfill
  \caption {The  architecture of the commonsense validation and reasoning with zero-shot prompting of LLMs.}
  \label{model architecture}
\end{figure*}

\section{Methodology}
Our study aims at evaluating the performance of multiple Large Language Models (LLMs) for commonsense validation and reasoning using zero-shot prompting. This approach leverages pre-trained LLMs without task-specific fine-tuning, relying solely on their inherent reasoning capabilities. For this purposes, we utilize the SemEval-2020 Task 4 dataset \cite{wang2020semeval}, which comprises labeled statements designed for commonsense validation and explanation tasks. To ensure a fair comparison between explicitly fine-tuned models and those evaluated solely with zero-shot prompting, we use only the test set for evaluation. The test set contains 1,000 entries for each task (Task A and Task B), providing a standardized benchmark for assessing model performance. The dataset is publicly available and can be accessed at \footnote{\url{https://github.com/wangcunxiang/SemEval2020-Task4-Commonsense-Validation-and-Explanation}}.

As depicted in Figure~\ref{model architecture}, the methodology consists of the following key stages:

\begin{itemize}
    \item \textbf{Pre-processing:} preparing the input test data templatic prompt to ensure compatibility with zero-shot prompting.
    \item \textbf{Model Calling:} Applying zero-shot prompting to multiple LLMs, including LLaMA3, Gemma2, and Mixtral to assess their commonsense validation and reasoning abilities. LLMs are directly accessible through the GroqCloud \footnote{\url{https://console.groq.com/docs/models}} Models API endpoint using the model IDs
    \item \textbf{Performance Metrics:} Evaluating model outputs based on accuracy to quantify their effectiveness.
    \item \textbf{Comparative Analysis:} Benchmarking zero-shot LLMs performance against fine-tuned transformer models to examine the impact of training on commonsense validation and reasoning tasks.
\end{itemize}

\begin{table*}[h]
\centering
\begin{tabular}{lcc}
\hline
\textbf{Model} & \textbf{Task A (Validation) (\%)} & \textbf{Task B (Explanation) (\%)} \\
\hline
Human & 99.1 & 97.8 \\
\hline
CN-HIT-IT.NLP & 97.0 & \textbf{94.80} \\
ECNU-SenseMaker & 96.7 & 95.0 \\
IIE-NLP-NUT & 96.4 & 94.3 \\
Solomon & 96.0 & 94.0 \\
\hline
L3-70B (LLaMA3-70B) & \textbf{98.40} & 93.40 \\
G2-9B (Gemma2-9B) & 97.90 & 91.00 \\
L3-8B (LLaMA3-8B) & 84.40 & 83.10 \\
M8x7B (Mixtral-8x7B) & 66.00 & 50.90 \\
\hline
\end{tabular}
\caption{Performance of different models on Task A (Commonsense Validation) and Task B (Commonsense Explanation) for English data. The models are: L3-70B (LLaMA3-70B), G2-9B (Gemma2-9B), L3-8B (LLaMA3-8B), and M8x7B (Mixtral-8x7B).}
\label{tab:taskAB-results}
\end{table*}

\begin{table*}
\centering
\begin{tabular}{r p{4cm} p{4cm} c c c c}
\hline
id & sent0 & sent1 & L3-70B & G2-9B & M8x7B & L3-8B \\
\hline
459 & The dog pounced on the rabbit & The cat pounced on the rabbit & sent0 & sent0 & \textbf{Other} & sent0 \\
737 & She purchased four supermarket tickets. & She purchased four theater tickets. & sent1 & sent1 & sent1 & sent1 \\
174 & Witches are not made of wood & Toads are not made of wood & sent0 & sent0 & sent0 & sent0 \\
\hline
\end{tabular}
\caption{Sample of common misclassified instances for TaskA. Model abbreviations: L3-70B = LLaMA3-70B, G2-9B = Gemma2-9B, M8x7B = Mixtral-8x7B, L3-8B = LLaMA3-8B. Keep in mind that Task A is about identifying which statement is against common sense?}
\label{tab:common_misclassified_taskA}
\end{table*}

\section{Results and Discussion}

Table \ref{tab:taskAB-results} presents the performance of the models on the commonsense validation (Task A) and commonsense explanation (Task B) tasks from SemEval-2020 Task 4. The results for human performance and transformer-based models (CN-HIT-IT.NLP, ECNU-SenseMaker, IIE-NLP-NUT, and Solomon) are as reported in the original SemEval-2020 Task 4 paper \cite{wang2020semeval}. In contrast, the results for the LLMs (LLaMA3, Gemma2, and Mixtral) are obtained from our experiments with zero-shot prompting. Findings are reported in the following subsections.

\subsection{Performance Analysis}

Among the models evaluated in this study, \textbf{L3-70B (LLaMA3-70B)} demonstrated the highest performance in Task A, scoring \textbf{98.4\%}, with an evidence that large-scale LLMs can effectively validate commonsense knowledge with zero-shot prompting. However, its performance in Task B (\textbf{93.4\%}) lags behind the transformer-based models reported as top 4 performing models in the Task paper. These models were explicitly fine-tuned for the task and some of them used external resources for the models training. This indicates that while LLMs are highly proficient in identifying implausible statements, they still struggle with selecting the most relevant explanation, demonstrating limitations in causal and inferential reasoning.

Similarly, the \textbf{G2-9B (Gemma2-9B)} model achieves strong performance in Task A (\textbf{97.9\%}) but showing a more significant decline in Task B (\textbf{91.0\%}). This further highlights the challenge of explanation selection, as these models may recognize implausibility without fully understanding the underlying causal mechanisms.

A size-dependent trend is observed in the LLaMA3 models. The smaller \textbf{L3-8B (LLaMA3-8B)} demonstrates significantly weaker performance than its larger version, with \textbf{84.4\%} in Task A and \textbf{83.1\%} in Task B. Finaly, the \textbf{M8x7B (Mixtral-8x7B)} model exhibited the weakest performance, with \textbf{66.0\%} in Task A and \textbf{50.9\%} in Task B. Its near-random performance in explanation selection suggests that it struggles not only with causal inference but also with general commonsense understanding, likely due to limitations in its training data or architecture. It is important to note that this lower accuracy was not due to weak reasoning abilities but rather due to inconsistencies in the output format, where the model provided both classification and explanation instead of following the expected template for the output.

\subsection{Implications for Zero-Shot Commonsense Reasoning}

The results indicate that while LLMs often recognize implausible statements but fail to select the most relevant explanation, highlighting deficits in causal and inferential reasoning. This suggests that current zero-shot approaches may capture surface-level plausibility but lack deeper reasoning abilities necessary for explanation generation.

Furthermore, the comparison between pre-trained LLMs and task-specific models from SemEval-2020 Task 4 suggests that explicit fine-tuning on commonsense explanation data remains beneficial. While larger models such as \textbf{L3-70B} outperform fine-tuned models in validation, they do not surpass them in explanation selection, reinforcing the need for additional adaptation to improve causal reasoning.

\begin{table*}
\centering
\begin{tabular}{r p{2cm} p{2cm} p{2cm} p{2cm} c c c c}
\hline
id & FalseSent. & OptionA & OptionB & OptionC & L3-70B & G2-9B & M8x7B & L3-8B \\
\hline
1388 & Roberts' room is sleeping & A room cannot close his eyes, because he has no eyes. & Robert won't let the room sleep because he needs rest. & Robert can sleep in his room & A & A & A & A \\
1444 & There are four years each season. & Different seasons have different temperatures. & A year can be divided into four seasons. & A season is shorter than a year. & C & C & \textbf{Other} & C \\
1172 & People can need sleep. & Sleep is not a thing to have it granted. & Sleeping is nature for every living being. & Sleeping is an activity that every living thing does. & A & A & C & A \\
\hline
\end{tabular}
\caption{Sample of common misclassified instances for TaskB. Model abbreviations: L3-70B = LLaMA3-70B, G2-9B = Gemma2-9B, M8x7B = Mixtral-8x7B, L3-8B = LLaMA3-8B. Task B is about selecting the reason for Why is this statement against common sense?}
\label{tab:common_misclassified_taskB}
\end{table*}

\subsection{Common Misclassification Patterns}

An analysis of misclassified instances provides insights into the reasoning patterns of different models. In \textbf{Task A}, some models failed to differentiate between subtle variations in sentence structure. For example, the model incorrectly classified the following pair:

\begin{quote}
\textit{The dog pounced on the rabbit.} \
\textit{The cat pounced on the rabbit.}
\end{quote}

This type of error suggests that the models may rely on statistical patterns rather than deep semantic understanding. 

In \textbf{Task B}, errors were primarily related to the selection of the most plausible explanation. A notable example is:

\begin{quote}
\textbf{False Statement:} "There are four years each season."\\
\textbf{Correct Explanation:} "A year can be divided into four seasons."
\end{quote}

Some models selected incorrect explanations, indicating potential limitations in their ability to link cause-effect relationships effectively. It should be noted that sentence IDs \textbf{1388, 1444, and 1172} are not present in the common misclassified instances of Task A. 

Despite the overall strong performance, the results also highlight challenges in certain reasoning aspects. The models demonstrated difficulty in selecting the most appropriate explanation for an implausible statement in Task B, even though they performed well in identifying implausible statements in Task A. This suggests that while the models recognize commonsense inconsistencies, they may struggle to justify their choices accurately. One possible explanation for this challenge is that Task B requires models to establish causal or inferential relationships between a false statement and its explanation. While Task A is a binary classification task requiring identification of implausible statements, Task B introduces additional complexity by demanding a deeper understanding of reasoning patterns and cause-effect relationships. Selecting the correct explanation requires not only recognizing a logical inconsistency but also evaluating multiple plausible justifications and determining which one best aligns with human commonsense knowledge. This suggests that current LLMs, despite their powerful language modeling capabilities, may still struggle with selecting the most contextually relevant explanation among multiple plausible options, as this task requires a nuanced understanding of real-world implications and reasoning structures \cite{mondorf2024beyond}.

Additionally, the low measured performance of \textbf{Mixtral-8x7B} can be attributed to output inconsistencies. The model frequently produced both an answer and an explanation, which deviated from the required response format. This indicates that we cannot rely  on the achieved results for this model to evaluate its performance on both tasks. More post-processing steps are required to ensure consistent output formatting when evaluating model performance.

\subsection{Conclusion}

This study demonstrates that large-scale LLMs, particularly \textbf{LLaMA3-70B} and \textbf{Gemma2-9B}, exhibit strong commonsense reasoning capabilities even in a zero-shot setting. These models outperform state-of-the-art fine-tuned transformer-based models, indicating that LLMs can generalize well across commonsense validation tasks. However, challenges remain in explanation selection and maintaining consistent output formats. Future research may include exploring Commonsense knowledge-graph LLMs \cite{li2021systematic, zhao2024large, toroghi2024right}, in addition to fine-tuning strategies, retrieval-augmented approaches, and structured prompting techniques to enhance the inferential reasoning capabilities of LLMs in zero-shot settings.

\bibliography{custom}

\end{document}